\documentclass[runningheads]{llncs}

\usepackage[T1]{fontenc}    
\usepackage{amsmath}
\usepackage{hyperref}       
\usepackage{url}            
\usepackage{booktabs}       
\usepackage{amsfonts}       
\usepackage{nicefrac}       
\usepackage{microtype}     
\usepackage{hyperref}
\usepackage{graphicx}
\usepackage{subcaption}
\usepackage{svg}
\usepackage{array}
\usepackage{multirow}
\usepackage{booktabs}
\usepackage{colortbl}
\usepackage{caption}   
\usepackage{subcaption} 
\usepackage{indentfirst}
\PassOptionsToPackage{table}{xcolor}
\usepackage{xcolor}
\newcommand{\hl}[1]{\cellcolor{blue!20}#1}
\usepackage{amssymb} 
\usepackage{marvosym}

\title{Forget-MI: Machine Unlearning for Forgetting Multimodal Information in Healthcare Settings}
\titlerunning{Forget-MI: Multimodal Unlearning in Healthcare}

\begin{document}

\author{
Shahad Hardan\textsuperscript{\textsection}\inst{1}\textsuperscript{\Letter} \and
Darya Taratynova\textsuperscript{\textsection}\inst{1} \and
Abdelmajid Essofi\inst{1} \and 
Karthik Nandakumar\inst{2,3} \and 
Mohammad Yaqub\inst{2}
}
\institute{%
Department of Machine Learning, Mohamed bin Zayed University of Artificial Intelligence, Abu Dhabi, UAE  \and
Department of Computer Vision, Mohamed bin Zayed University of Artificial Intelligence, Abu Dhabi, UAE  \and
Department of Computer Science, Michigan State University, United States \\
\vspace{0.2cm}
\email{shahad.hardan@mbzuai.ac.ae, darya.taratynova@mbzuai.ac.ae, abdelmajid.essofi@mbzuai.ac.ae, nandakum@msu.edu, mohammad.yaqub@mbzuai.ac.ae}
}
\authorrunning{S. Hardan et al.}

\maketitle

\begingroup
\renewcommand\thefootnote{\textsection}
\footnotetext{These authors contributed equally to this work.}
\renewcommand\thefootnote{\Letter}
\footnotetext{Corresponding author: \email{shahad.hardan@mbzuai.ac.ae}}
\endgroup

\begin{abstract}
Privacy preservation in AI is crucial, especially in healthcare, where models rely on sensitive patient data. In the emerging field of machine unlearning, existing methodologies struggle to remove patient data from trained multimodal architectures, which are widely used in healthcare. We propose \textit{Forget-MI}, a novel machine unlearning method for multimodal medical data, by establishing loss functions and perturbation techniques. Our approach unlearns unimodal and joint representations of the data requested to be forgotten while preserving knowledge from the remaining data and maintaining comparable performance to the original model. We evaluate our results using performance on the forget dataset $\downarrow$, performance on the test dataset $\uparrow$, and Membership Inference Attack (MIA) $\downarrow$, which measures the attacker's ability to distinguish the forget dataset from the training dataset. Our model outperforms the existing approaches that aim to reduce MIA and the performance on the forget dataset while keeping an equivalent performance on the test set. Specifically, our approach reduces MIA by $0.202$ and decreases AUC and F1 scores on the forget set by $0.221$ and $0.305$, respectively. Additionally, our performance on the test set matches that of the retrained model, while allowing forgetting. Code is available at \url{https://github.com/BioMedIA-MBZUAI/Forget-MI.git}.
\keywords{Machine Unlearning \and Multimodal Data \and Clinical Data \and Privacy Preserving \and Right to Be Forgotten }
\end{abstract}

\section{Introduction}

\indent The wide and fast-paced applications of artificial intelligence (AI) may, in many cases, necessitate the use of private data. This usage threatens the privacy and security of individuals or organizations. To address such concerns, the ``right to be forgotten'' is legislated in many countries around the world \cite{us_regulation,gdpr,canada_regulation}. This requirement aims to guarantee the removal of individual data upon request as well as its effect on model learning, leading to the emergence of the field of machine unlearning. In healthcare settings, the use of medical data in AI models is highly restricted due to its sensitivity and confidentiality. Therefore, machine unlearning can contribute significantly to addressing patient privacy concerns and the robustness of de-identification methods \cite{machine_unlearning_survey}. For instance, machine unlearning can assure the removal of patient data and fulfill withdrawal requests from subjects involved in clinical research \cite{biobank_withdrawal}. 

\begin{figure}[t]
    \centering
    \includegraphics[scale=0.14 ]{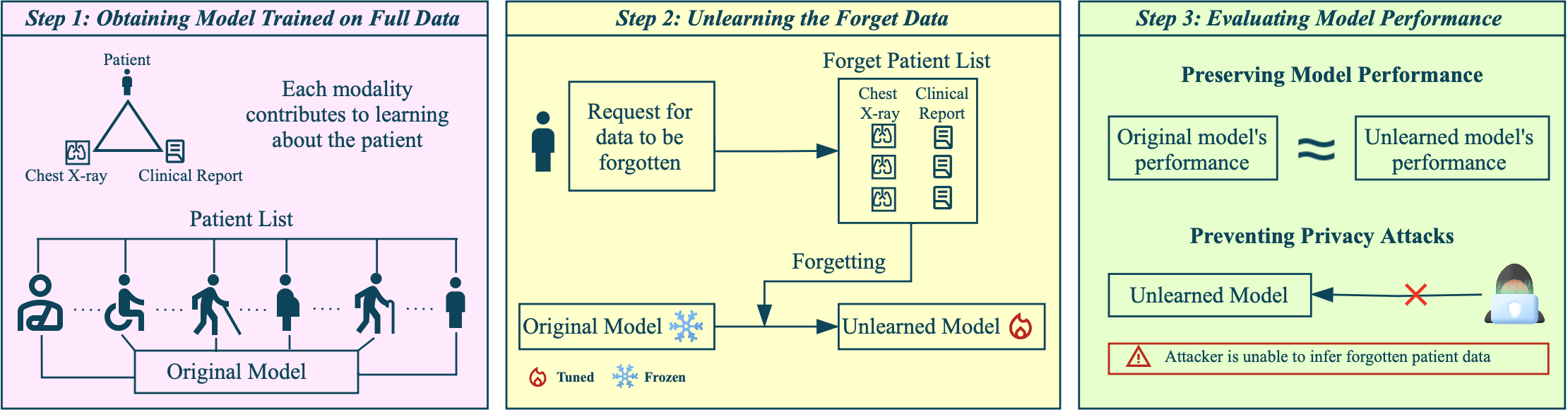}
    \caption{The original setting contains a model trained on multimodal patient data. When a patient requests their data to be forgotten, we create the forget dataset. This dataset is then forgotten from the original model, obtaining the unlearned model. The unlearned model 1) prevents access to the forgotten data and 2) preserves the original model's performance on the retained data.}
    \label{fig:unlearning_general}
\end{figure}

The goal of machine unlearning is to ensure that the unlearned model achieves a precision comparable to that of the original model trained on the full dataset \cite{sisa}, while forgetting the data requested for removal. The process of machine unlearning is illustrated in Fig. \ref{fig:unlearning_general}. Primarily, we have 1) a dataset containing patient data of different modalities that collectively contribute to learning about the patient and 2) an original model trained on the full patient dataset (Step 1, Fig. \ref{fig:unlearning_general}). When a patient requests the removal of their data from the model, we create the forget dataset containing all used modalities. The patient's data is forgotten by unlearning it from the original model, from which we obtain the unlearned model (Step 2, Fig. \ref{fig:unlearning_general}). To evaluate the effectiveness of unlearning, we have to ensure that 1) the unlearned model's performance is close enough to the original model's performance and 2) any attack to retrieve forgotten data from the unlearned model is unsuccessful (Step 3, Fig. \ref{fig:unlearning_general}).

However, it is challenging to confirm the erasure of the forget dataset due to the lack of explainability of most architectures and the randomness in models. One method, known as exact unlearning, involves retraining the model from scratch while excluding the forget data \cite{7163042,exact2,exact1}. However, this comes at a disadvantage because 1) retraining the model is expensive and time-consuming and 2) there is no guarantee to still have access to the training dataset. On the other hand, approximate unlearning efficiently modifies the existing model to avoid the drawbacks of retraining \cite{machine_unlearning_survey}. Furthermore, medical data is collected from multiple sources and in different types, such as images and clinical notes \cite{multimodal_ai,fusion_healthcare}. Thus, the fusion of these modalities, a common practice, motivates the need for multimodal unlearning techniques to address the complexities of multimodal approaches. Combining these valuable aspects, our study explores machine unlearning within the context of multimodal medical data. 

Among the notable frameworks used for unlearning, teacher-student frameworks have emerged as an effective approach, in which the student model is guided by the teacher through distillation or regularization, ensuring that the student retains overall performance while unlearning the specific data points to be forgotten \cite{badteaching,scrub,Zhou_2023}. Other machine unlearning methods identify and update only the parameters most affected by the forget set \cite{amnesiacmu,hessians,shi2024deepcleanmachineunlearningcheap}.

In multimodal settings, prior machine unlearning methods focused on modality decoupling in image-text systems \cite{multidelete,lipschitz,kravets2024zeroshotclassunlearningclip,li2024singleimageunlearningefficient}. Our work is partly inspired by \cite{multidelete}, which proposed breaking associations between modality pairs marked for deletion while preserving unimodal knowledge. However, this unimodal retention is problematic in healthcare settings. Therefore, a new method must be designed to erase both unimodal and multimodal patient information. Accordingly, we summarize our contributions as follows:
\begin{itemize}
    \item We introduce, \textit{Forget-MI}, a novel method capable of efficiently unlearning unimodal and joint embeddings from multimodal architectures.
    \item We are the first to introduce multimodal unlearning in healthcare where we account for the multiple studies per patient in medical datasets.
    \item  We propose a system of loss functions that incorporate noise to help control the distance between the unlearning and original model embeddings to tackle challenges of narrow embedding distributions in medical data. 
    \item We evaluate \textit{Forget-MI} by testing its robustness against attacks, ensuring privacy protection while largely preserving performance on retained data across different unlearning levels.
\end{itemize}

\section{Methodology}

\noindent\textbf{Problem Formulation:} We introduce, \textit{Forget-MI}, a novel method that unlearns joint and unimodal representations, ensuring patient data is forgotten from the trained model. Given an original model $\mathcal{F}_{\textbf{og}}$ trained on the full dataset $D$, we aim to unlearn a subset of data points, resulting in an unlearned model $\mathcal{F}_{\textbf{ul}}$ that forgets this information while retaining the remaining knowledge from the original model. In order to start unlearning from $\mathcal{F}_{\textbf{og}}$, we extract the set of points we want to unlearn, forming the forget set $D_f \in D$, then the remaining datapoints form the retain set $D_r = D - D_f$. Starting with $\mathcal{F}_{\textbf{ul}} = \mathcal{F}_{\textbf{og}}$, we aim to lead our desired model $\mathcal{F}_{\textbf{ul}}$ to perform on $D_f$ similar to the way that the original model $\mathcal{F}_{\textbf{og}}$ would perform on unseen data.
\begin{figure}[tbh]
    \centering
    \includegraphics[scale=0.1]{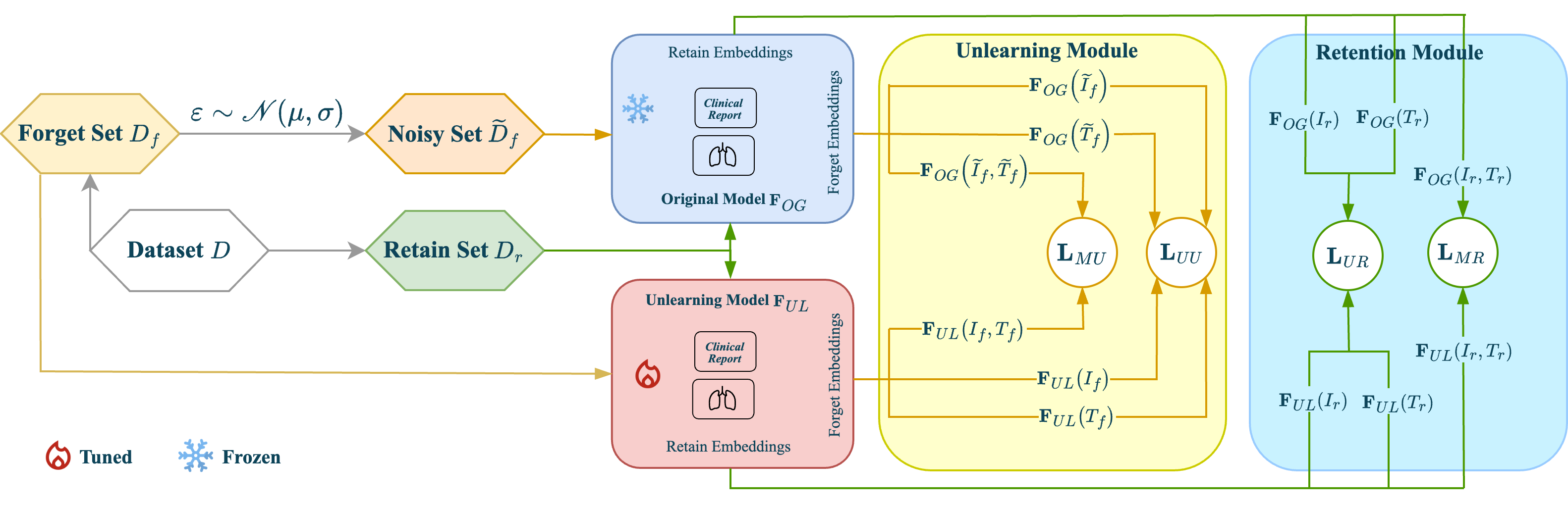}
    \caption{\textit{Forget-MI}'s Approach: The dataset has images $I$ and text $T$, and is split into retain $D_r$ and forget $D_f$, with noise added to the $D_f$ to create $\tilde{D}_f$. We have two models: original $\mathcal{F}_{\textbf{og}}$ and unlearning $\mathcal{F}_{\textbf{ul}}$, and we start with $\mathcal{F}_{\textbf{ul}}=\mathcal{F}_{\textbf{og}}$. Then, $\mathcal{F}_{\textbf{ul}}$ unlearns by computing the total loss that consists of: unimodal unlearning $\mathcal{L}_{UU}$, multimodal unlearning $\mathcal{L}_{MU}$, unimodal retention $\mathcal{L}_{UR}$, and multimodal retention $\mathcal{L}_{MR}$. The unlearning losses induce forgetting $D_f$, while the retention losses preserve the knowledge on $D_r$.}
    \label{fig:unlearning_diagram}
\end{figure}

Our unlearning process adjusts the model's loss function in a way that promotes forgetting $D_f$ while preserving the information in $D_r$, as described in Fig. \ref{fig:unlearning_diagram}. We achieve this by computing four losses and backpropagating their sum. The losses are defined below, where $[\mathcal{F}(T), \mathcal{F}(I)]$ is a concatenation of the unimodal embeddings produced by a model $\mathcal{F}$, and $\mathcal{F}(T, I)$ is the joint embedding: \\
\textbf{Unimodal Unlearning (UU):} This loss targets forgetting the unimodal encodings within the forget set $D_f$. We enable forgetting a neighborhood around the image-text pair $(I_f, T_f) \in D_f$ by adding noise to form $(\tilde{I_f}, \tilde{T_f})$. So, the UU loss pushes $\mathcal{F}_{\textbf{ul}}$'s unimodal embeddings on $(I_f, T_f)$ away from $\mathcal{F}_{\textbf{og}}$'s unimodal embeddings on $(\tilde{I_f}, \tilde{T_f})$. This distance is computed as follows:
\begin{align}
    \mathcal{L}_{UU} = - Dist\Big([\mathcal{F}_{\textbf{ul}}(I_f), \mathcal{F}_{\textbf{ul}}(T_f)], [\mathcal{F}_{\textbf{og}}(\tilde{I_f}), \mathcal{F}_{\textbf{og}}(\tilde{T_f})]\Big).
    \label{eq:loss_uu}
\end{align}
\textbf{Multimodal Unlearning (MU):} This loss targets the edges between modalities in the forget set $D_f$, by achieving the same functionality of the UU loss but on the joint embeddings to break the correlation between modalities. We set $(I_f, T_f) \in D_f$ as the modality pair to be unlearned from $\mathcal{F}_{\textbf{og}}$, and $(\tilde{I_f}, \tilde{T_f})$ as its noisy counterpart. To encourage the joint embedding of $\mathcal{F}_{\textbf{ul}}$ to dissociate from the joint embedding of $\mathcal{F}_{\textbf{og}}$, we define the MU loss by the following formula:
\begin{align}
    \mathcal{L}_{MU} = - Dist\Big(\mathcal{F}_{\textbf{ul}}((I_f, T_f)), \mathcal{F}_{\textbf{og}}((\tilde{I_f}, \tilde{T_f}))\Big).
    \label{eq:loss_mu}
\end{align}
\textbf{Unimodal Retention (UR):} While unlearning $D_f$, some information from $D_r$ may also be forgotten, resulting in unintended information loss. To tackle that, UR loss helps the model maintain its performance on the image and text unimodal representation of the retain set. We set $(I_r, T_r) \in D_r$ as the modality pair in the retain set. We want the unimodal encodings in $\mathcal{F}_{\textbf{og}}$ to be preserved in $\mathcal{F}_{\textbf{ul}}$, so we define the UR loss by the following formula:
\begin{align}
    \mathcal{L}_{UR} = Dist\Big([\mathcal{F}_{\textbf{ul}}(I_r), \mathcal{F}_{\textbf{ul}}(T_r)], [\mathcal{F}_{\textbf{og}}(I_r), \mathcal{F}_{\textbf{og}}(T_r)]\Big).
    \label{eq:loss_ukr}
\end{align}
\textbf{Multimodal Retention (MR):} Following the UR loss, we retain the joint embeddings in the unlearned model $\mathcal{F}_{\textbf{ul}}$. We set $(I_r, T_r) \in D_r$ as the modality pair from the retain set. We want $\mathcal{F}_{\textbf{ul}}$ to maintain the same precision on $(I_r, T_r)$ as $\mathcal{F}_{\textbf{og}}$, so we define the MR loss by the following formula:
\begin{align}
    \mathcal{L}_{MR} = Dist\Big(\mathcal{F}_{\textbf{ul}}((I_r, T_r)), \mathcal{F}_{\textbf{og}}((I_r, T_r))\Big).
    \label{eq:loss_mkr}
\end{align}  
\textbf{Total Loss:} These losses formulate our objective of unlearning the data in $D_f$ while retaining the desired information in $D_r$. Unlearning is achieved by backpropagating on the total loss as follows: 
\begin{align}
    \mathcal{L} = w_{uu} \mathcal{L}_{UU} + w_{ur}
 \mathcal{L}_{UR} + w_{mu} \mathcal{L}_{MU} + w_{mr}
 \mathcal{L}_{MR},
\end{align}
where \( w_{uu} + w_{ur} + w_{mu} + w_{mr} = 1 \).

\section{Experimental Setup}

\noindent \textbf{Dataset:} We conduct experiments on the MIMIC Chest X-ray (MIMIC-CXR) database \cite{mimic-cxr}. We work on a subset of MIMIC-CXR, adopted from \cite{joint_cxr}, which contains 6,742 images and radiology reports belonging to 1,663 subjects. Our data is labeled according to the four stages of edema: no edema (43\%), vascular congestion (25\%), interstitial edema (22\%), and alveolar edema (10\%), leading to a multi-classification problem. \\
\textbf{Implementation Details:} Since access to the full training dataset is not always feasible, explicitly defining $D_r =D - D_f$ is problematic. Accordingly, in our experiments, the size of the retain set is fixed to match that of the forget set at each epoch, and a subset is constructed by sequentially sampling points from the available retained data. Unlearning begins with an original model that has been trained on the full dataset. We adopt from \cite{joint_cxr} a multimodal architecture that uses late fusion and encodes CXR images using ResNet and radiology reports (text) using a pre-trained SciBERT model \cite{scibert}. Furthermore, for the model to forget a neighborhood of the desired data points, we apply noise to the forget set, as denoted in Equations \eqref{eq:loss_uu} and \eqref{eq:loss_mu}. We handle noise on the image and text levels differently. For images, we apply Gaussian noise \cite{lipschitz} with hyperparameters $\mu$ and $\sigma $, representing mean and standard deviation of the noise distribution. We experiment on a set of $\mu = \{0, 0.1, 0.2\}$ and $\sigma = \{0.1, 0.2\}$. For text, we add noise at a character level and a word level randomly \cite{syn-text}. After experimenting with different levels, we observe that the noise has to be small to avoid distorting the model's performance.

In addition, we construct joint embeddings for unlearning using a multimodal adaptation gate that represents the two modalities while maintaining the visual as the dominant one \cite{gate}. There are two learning rates used in the unlearning experiments: $1e-4$ and $1e-5$, for 30 epochs. The distance used in Equations \eqref{eq:loss_uu}-\eqref{eq:loss_mkr} is the Euclidean distance.\\
\textbf{Forming Forget Datasets:}  In a healthcare setting, we address cases where patients request the removal of \textit{all} their data, which may encompass multiple studies per patient. Our forget sets include 3\%, 6\%, and 10\% of the original dataset size, aiming to study the effect of the forget set size on the performance of the unlearning model. To ensure fair sample selection, the forget datasets preserve the overall distribution of study counts per patient, including both patients with multiple studies and those with fewer studies. \\
\textbf{Evaluation}: Following previous works \cite{multidelete,badteaching,lipschitz}, we employ standard metrics to evaluate the unlearning process.
The first metric is $\textbf{distance} \downarrow$, which is inspired by differential privacy to quantify the difference between the unlearned and original models on the retain and test samples. The second metric is the Membership Inference Attack (\textbf{MIA} $\downarrow$) in a black-box setting, where we only have access to the model outputs. Following \cite{scrub}, we train an SVM classifier with losses from the retain samples as positive examples and losses from the test samples as negative examples. Finally, we evaluate the unlearned model’s performance using \textbf{macro-F1} score and \textbf{AUC} on two datasets: the forget set ($D_f$ $\downarrow$) and the test set ($D_t$ $\uparrow$). 

\section{Results \& Disucssions} \label{sec:results}

We compare \textit{Forget-MI} with modality-agnostic approaches, such as NegGrad+ \cite{scrub}, which fine-tunes the original model on \(D_r\) while negating the gradient for \(D_f\); SCRUB \cite{scrub}, a teacher-student approach where the student model is trained to selectively disobey the teacher on \(D_f\) while aligning on \(D_r\); Catastrophic Forgetting (CF-\(k\)) \cite{euk}, where the first $k =2$ layers of the original model are frozen and the remaining layers are fine-tuned on \(D_r\); and Exact Unlearning (EU-\textit{\(k\))} \cite{euk}, where the unfrozen layers are retrained on \(D_r\). From a  multimodal perspective, we evaluate MultiDelete \cite{multidelete}, which retains multimodal and unimodal information on \(D_r\) while only unlearning $D_f$ on a multimodal level.

Aside from the baselines, we compare \textit{Forget-MI} with a retrained model. We conduct experiments where we explore the effect of different noise and weight settings on the unlearning results. Our experimental settings include: 1) \textit{\hyperref[tab_fm]{No Noise}}, adding no distortion to the forget set, 2) \textit{\hyperref[tab_fm]{Equal}} weights for all losses, 3) higher weights for \textit{\hyperref[tab_fm]{Multimodal}} losses $w_{mu}$ and $w_{mr}$, 4) higher weights for \textit{\hyperref[tab_fm]{Unimodal}} losses $w_{uu}$ and $w_{ur}$, and 5) higher weights for \textit{\hyperref[tab_fm]{Retention}} losses  $w_{ur}$ and $w_{mr}$. This methodology is applied consistently across forget set sizes of 3\%, 6\%, and 10\% of the data, with the Gaussian noise having $\mu = 0$ and $\sigma = 0.1$. We compare our best \textit{Forget-MI} setting for each forgetting percentage with the retrained model and the baselines in Table \ref{tab:baseline_comp}.  Additionally, we compare the results of different \textit{Forget-MI} settings with the original and retrained models in Table \ref{tab_fm}.
\begin{table}[t]
\centering
\caption{Comparison between our best \textit{Forget-MI} setting, the retrained model, and the baseline models on forgetting 3\%, 6\%, and 10\% of the data. The Table shows that \textit{Forget-MI} is closer to the retrained model in all cases. The \textit{Forget-MI} experiment used is the best loss weightage setting for each of the percentages, which is then detailed in Table \ref{tab_fm}. The highlighted results show the best \textbf{combination} of metrics.}
\resizebox{\textwidth}{!}{
\begin{tabular}{l|ccccc|ccccc|ccccc}
\toprule
\multirow{2}{*}{Method} & \multicolumn{5}{c|}{3\%} & \multicolumn{5}{c|}{6\%} & \multicolumn{5}{c}{10\%} \\
\cmidrule(lr){2-6} \cmidrule(lr){7-11} \cmidrule(lr){12-16}
& MIA $\downarrow$ & $D_f,_\text{AUC} \downarrow$ & $D_f,_\text{F1} \downarrow$ & $D_t,_\text{AUC} \uparrow$ & $D_t,_\text{F1} \uparrow$ 
& MIA $\downarrow$ & $D_f,_\text{AUC} \downarrow$ & $D_f,_\text{F1} \downarrow$ & $D_t,_\text{AUC} \uparrow$ & $D_t,_\text{F1} \uparrow$ 
& MIA $\downarrow$ & $D_f,_\text{AUC} \downarrow$ & $D_f,_\text{F1} \downarrow$ & $D_t,_\text{AUC} \uparrow$ & $D_t,_\text{F1} \uparrow$ \\
\midrule
Original & 1.000 & 0.999 & 0.965 & 0.677 & 0.388
                              & 1.000 & 0.999 & 0.972 & 0.677 & 0.388
                              & 1.000 & 0.999 & 0.970 & 0.677 & 0.388 \\
Retrain & 0.000 & 0.566 & 0.310 & 0.626 & 0.362
        & 0.769 & 0.675 & 0.395 & 0.702 & 0.427 
        & 0.190 & 0.588 & 0.342 & 0.629 & 0.382 \\ 
\midrule
NegGrad+ & 1.000 & 1.000 & 0.983 & 0.678 & 0.399
        & 1.000 & 1.000 &0.987 & 0.677 & 0.388 
        & 1.000 & 1.000 & 0.986 & 0.677 & 0.388 \\ 

SCRUB & 1.000 & 1.000 & 0.983 & 0.679 & 0.397 
        & 1.000 & 1.000 & 0.987 & 0.679 & 0.394 
        & 1.000 & 1.000 & 0.985 & 0.679 & 0.394 \\

CF-k  & 1.000 & 0.999 & 0.976 & 0.686 & 0.416 
                & 1.000 & 0.999 & 0.977 & 0.676 & 0.410 
                & 1.000 & 1.000 & 0.990 & 0.668 & 0.410 \\

EU-k & 1.000 & 0.996 & 0.952 & 0.669 & 0.386 
                & 1.000 & 0.612 & 0.328 & 0.631 & 0.339
                & 1.000 & 0.629 & 0.345 & 0.631 & 0.339 \\

MultiDelete & 0.806 & 0.858 & 0.562 & 0.622 & 0.300 
              & 0.846 & 0.878 & 0.604 & 0.636 & 0.289 
              & 0.952 & 0.972 & 0.783 & 0.656 & 0.333 \\

\midrule
\textit{Forget-MI} & \hl{0.571} & \hl{0.735} & \hl{0.393} & \hl{0.625} & \hl{0.250}
     & \hl{0.615} & \hl{0.654} & \hl{0.328} & \hl{0.599} & \hl{0.270}                  
     & \hl{0.810} & \hl{0.656} & \hl{0.313} & \hl{0.565} & \hl{0.252} \\
\bottomrule
\end{tabular}}
\label{tab:baseline_comp}
\end{table}

\begin{table}[ht]
  \centering
  \caption{\textit{Forget-MI} on forgetting 3\%, 6\%, and 10\% of the data. Here, $D_f$ refers to the forget set and $D_t$ refers to the test set. The \textit{Forget-MI} experiments correspond to higher weighting to the respective losses (e.g., \textit{Forget-MI} Multimodal weights $w_{mu}$ and $w_{mr}$ more), as explained in Section \ref{sec:results}. The highlighted results show the best \textbf{combination} of metrics.}
  \resizebox{\textwidth}{!}{%
    \begin{tabular}{l|ccccc|ccccc|ccccc}
      \toprule
      \multirow{2}{*}{Method}
        & \multicolumn{5}{c|}{3\%}
        & \multicolumn{5}{c|}{6\%}
        & \multicolumn{5}{c}{10\%} \\
      \cmidrule(lr){2-6} \cmidrule(lr){7-11} \cmidrule(lr){12-16}
        & MIA $\downarrow$
        & $D_f,\text{AUC}\downarrow$
        & $D_f,\text{F1}\downarrow$
        & $D_t,\text{AUC}\uparrow$
        & $D_t,\text{F1}\uparrow$
        & MIA $\downarrow$
        & $D_f,\text{AUC}\downarrow$
        & $D_f,\text{F1}\downarrow$
        & $D_t,\text{AUC}\uparrow$
        & $D_t,\text{F1}\uparrow$
        & MIA $\downarrow$
        & $D_f,\text{AUC}\downarrow$
        & $D_f,\text{F1}\downarrow$
        & $D_t,\text{AUC}\uparrow$
        & $D_t,\text{F1}\uparrow$ \\
      \midrule
      Original
        & 1.000 & 0.999 & 0.965 & 0.677 & 0.388
        & 1.000 & 0.999 & 0.972 & 0.677 & 0.388
        & 1.000 & 0.999 & 0.970 & 0.677 & 0.388 \\
      Retrain
        & 0.000 & 0.566 & 0.310 & 0.626 & 0.362
        & 0.769 & 0.675 & 0.395 & 0.702 & 0.427
        & 0.190 & 0.588 & 0.342 & 0.629 & 0.382 \\
      \midrule
      No Noise
        & 1.000 & 0.534 & 0.116 & 0.508 & 0.154
        & 1.000 & 0.511 & 0.110 & 0.521 & 0.090
        & 1.000 & 0.513 & 0.104 & 0.505 & 0.085 \\
      Equal
        & 0.571 & 0.764 & 0.385 & 0.631 & 0.253
        & 0.846 & 0.802 & 0.428 & 0.637 & 0.277
        & 0.952 & 0.796 & 0.369 & 0.617 & 0.240 \\
      Multimodal
        & 0.714 & 0.774 & 0.384 & 0.634 & 0.247
        & \hl{0.615} & \hl{0.654} & \hl{0.328} & \hl{0.599} & \hl{0.270}
        & 0.905 & 0.740 & 0.344 & 0.586 & 0.231 \\
      Unimodal
        & \hl{0.571} & \hl{0.735} & \hl{0.393} & \hl{0.625} & \hl{0.250}
        & 0.615 & 0.687 & 0.357 & 0.610 & 0.275
        & 0.905 & 0.764 & 0.363 & 0.597 & 0.236 \\
      Retention
        & 0.714 & 0.646 & 0.306 & 0.598 & 0.245
        & 0.615 & 0.672 & 0.339 & 0.594 & 0.267
        & \hl{0.810} & \hl{0.656} & \hl{0.313} & \hl{0.565} & \hl{0.252} \\
      \bottomrule
    \end{tabular}%
  }
  \label{tab_fm}
\end{table}

\begin{figure}[h!]
    \centering
    \begin{subfigure}[b]{0.24\textwidth}
        \centering
        \includegraphics[width=\textwidth]{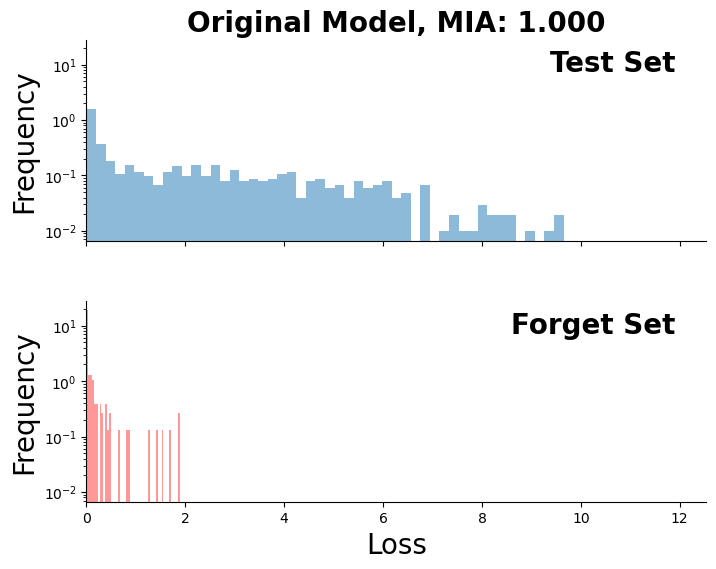}
        \caption{Original Model}
        \label{fig:original}
    \end{subfigure}
    \begin{subfigure}[b]{0.24\textwidth}
        \centering
        \includegraphics[width=\textwidth]{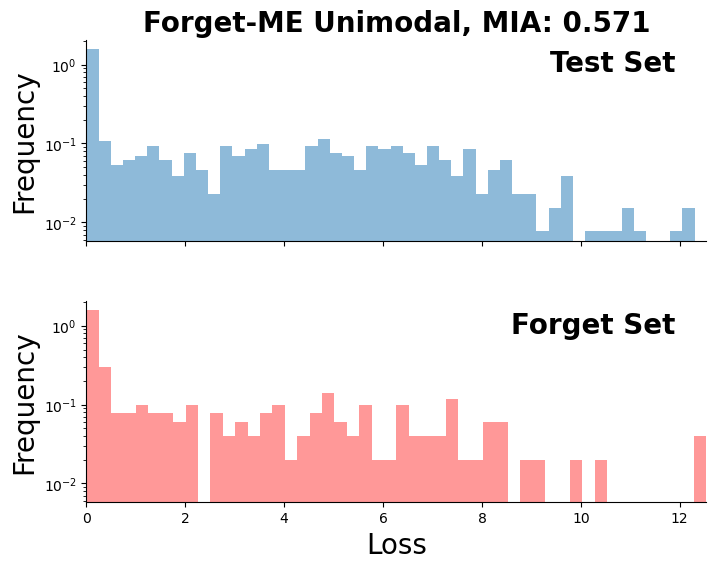}
        \caption{Forgetting 3\%}
        \label{fig:forget3}
    \end{subfigure}
    \begin{subfigure}[b]{0.24\textwidth}
        \centering
        \includegraphics[width=\textwidth]{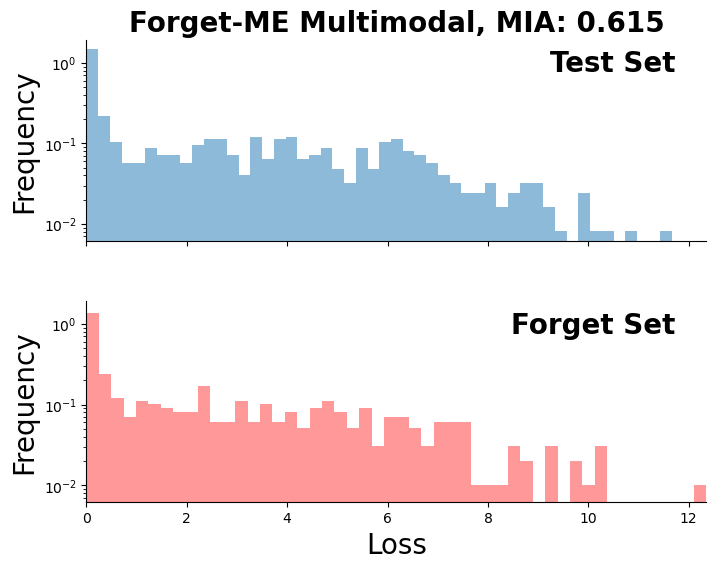}
        \caption{Forgetting 6\%}
        \label{fig:forget6}
    \end{subfigure}
    \begin{subfigure}[b]{0.24\textwidth}
        \centering
        \includegraphics[width=\textwidth]{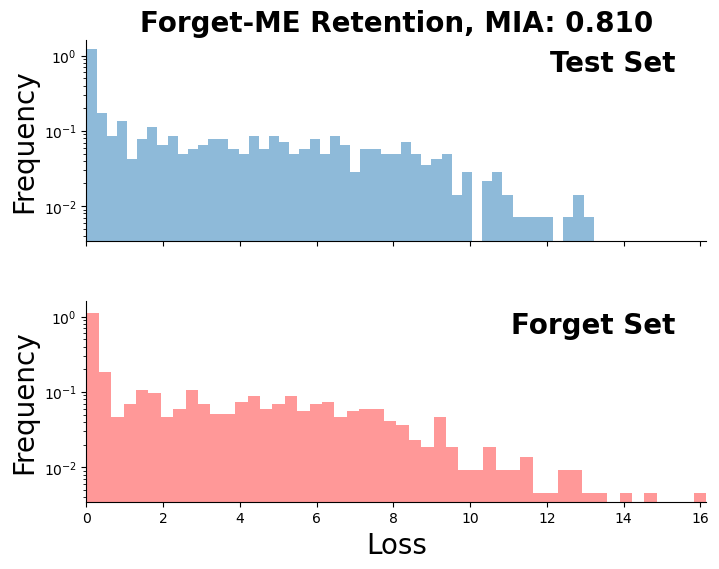}
        \caption{Forgetting 10\%}
        \label{fig:forget10}
    \end{subfigure}
    \caption{The original model vs. best \textit{Forget-MI} at each of forgetting percentages. The subplots shows the loss distribution for the forget set relative to the test set. The plots show that, unlike the original model, our model fails at recognizing the test set from the forget set, indicating successful unlearning.}
    \label{fig:all_histograms}
\end{figure}

Our experiments show that \textit{Forget-MI} consistently outperforms baseline methods, achieving the lowest MIA scores while maintaining comparable performance on the test set relative to the original and retrained models.\\
\textbf{Effect of Forget Percentage:} We identify the best \textit{Forget-MI} loss weighting settings for each forgetting percentage. When unlearning 3\% of the data, the \textit{Unimodal} setting yields the best results. At 6\%, where the MIA score is similar between \textit{Unimodal} and \textit{Multimodal} settings, we consider the \textit{Multimodal} experiment to be outperforming due to its lower performance on the forget set.  One reason behind this pattern is that the most distinguishing features in datasets with high inter-patient similarity are not found solely in individual records but are most uniquely captured in joint modality encodings. Analogously to humans, when our observations are abundant, our ability to remember and learn them increases when we connect our knowledge. On the other hand, when we unlearn 10\% of the data, it is best to focus more on the \textit{Retention} losses, as the model tends to compromise more on the performance on the test set due to its objective of forgetting more data. \\
\textbf{Evaluation Criterion:} Given the trade-off between unlearned model generalizability and the unlearning quality, identifying the optimal combination of metrics is more valuable than the conventional way of evaluation through the test set performance. The evaluation was based on the distance between the retrained model and the unlearning model, the MIA score, the performance on the forget set, and the performance on the test set. One way to demonstrate the efficiency of MIA-based unlearning assessment, according to \cite{unlearning_comp}, is to explore the loss distributions in the test and forget datasets of the original and each of the best unlearning models for the forgetting percentages, as shown in Fig. \ref{fig:all_histograms}. As illustrated, the original model can recognize the difference in the test and forget loss distributions. However, the unlearned models fail to do so, indicating that they no longer retain information from the forget set.\\
In addition, when observing very similar MIA scores for different configurations under the same forgetting percentage, we compare their performance on the forget and test sets. While we expect the model to struggle at evaluating the forget set, it should not come at the cost of low generalizability on the test set, making the model less useful after unlearning. \\
\textbf{Baseline Comparison:} We notice a clear difference between modality-agnostic and modality-aware approaches. Other than MultiDelete \cite{multidelete}, the baselines showed unsuccessful unlearning. This is explained by the fact that their average distance to the retrained model is $0.323$, while the original model's distance is $0.251$, indicating that modality-agnostic approaches do not push the model further away from the original. This is opposed to $0.483$, $0.410$, and $0.561$ for our best configurations when forgetting $3\%$, $6\%$ and $10\%$ of the data, respectively. On the other hand, MultiDelete \cite{multidelete} provides higher F1 than \textit{Forget-MI}. Due to the high data imbalance in this multi-classification problem, \textit{Forget-MI}'s superior ability to forget compromises on the performance on the low represented classes, causing a lower F1.\\
\textbf{Effect of noise:} We further demonstrated that introducing small Gaussian noise to the forget set significantly enhances the unlearning performance. Noiseless experiments resulted in the poorest outcomes due to the limited influence of isolated points on the broader data distribution. Thus, the added noise mitigates this by creating a neighborhood around forgotten data points, facilitating broader shifts in the representation of the model. Moreover, increasing the standard deviation consistently raises the MIA score, increases the forget set's performance, and reduces generalization. A higher mean decreases the MIA score and reduces the test set's generalization.\\
\textbf{Unlearning Time:} The computational cost of unlearning varies substantially across methods. Full retraining is the most time-intensive, requiring up to 14 hours. Among the unlearning approaches, NegGrad+, SCRUB, CF-k and EU-k complete within 4 hours. Forget-MI and MultiDelete require longer runtimes, taking up to 5 hours depending on the forget percentage. This trade-off in runtime is acceptable given the improved unlearning efficacy observed in these methods.

\section{Conclusion}

The robust performance of \textit{Forget-MI} represents  a significant step toward practical machine unlearning in sensitive domains like healthcare, effectively balancing unlearning performance, model utility, and privacy concerns. Generally, we observe that the model is effectively able to unlearn different percentages of data that we aim to forget, and it becomes more challenging as the percentage increases. This suggests inherent limitations in current unlearning algorithms' ability when faced with larger data removal requests. Moreover, future work in the healthcare field can extend to other clinical tasks and experiment with foundation models, which require high generalizability on unseen data, making forgetting more difficult.

\subsubsection{\discintname}
The authors have no competing interests to declare that are relevant to the content of this article.

\bibliographystyle{plain} 
\bibliography{reference} 

\end{document}